\documentclass{article} 
\usepackage{iclr2017_conference,times}
\usepackage[colorlinks=true]{hyperref}
\usepackage{url}
\usepackage{graphicx}
\usepackage{amsmath}
\usepackage{enumitem}

\title{Coupled Ensembles of Neural Networks}

\author{Anuvabh Dutt \\
Univ. Grenoble Alpes, CNRS, Grenoble-INP, LIG, F-38000 Grenoble France \\
\texttt{Anuvabh.Dutt@univ-grenoble-alpes.fr} \\
\And
Denis Pellerin \\
Univ. Grenoble Alpes, CNRS, Grenoble-INP, GIPSA-Lab, F-38000 Grenoble France \\
\texttt{Denis.Pellerin@gipsa-lab.grenoble-inp.fr} \\
\AND
Georges Qu\'enot \\
Univ. Grenoble Alpes, CNRS, Grenoble-INP, LIG, F-38000 Grenoble France \\
\texttt{Georges.Quenot@imag.fr}}

\begin{document}

\maketitle

\begin{abstract}
We investigate in this paper the architecture of deep convolutional networks. Building on existing state of the art models, we propose a reconfiguration of the model parameters into several parallel branches at the global network level, with each branch being a standalone CNN. We show that this arrangement is an efficient way to significantly reduce the number of parameters without losing performance or to significantly improve the performance with the same level of performance. The use of branches brings an additional form of regularization. In addition to the split into parallel branches, we propose a tighter coupling of these branches by placing the ``fuse (averaging) layer'' before the Log-Likelihood and SoftMax layers during training. This gives another significant performance improvement, the tighter coupling favouring the learning of better representations, even at the level of the individual branches. We refer to this branched architecture as ``coupled ensembles''. The approach is very generic and can be applied with almost any DCNN architecture. With coupled ensembles of DenseNet-BC and parameter budget of 25M, we obtain error rates of 2.92\%, 15.68\% and 1.50\% respectively on CIFAR-10, CIFAR-100 and SVHN tasks. For the same budget, DenseNet-BC has error rate of 3.46\%, 17.18\%, and 1.8\% respectively.  With ensembles of coupled ensembles, of DenseNet-BC networks, with 50M total parameters, we obtain error rates of 2.72\%, 15.13\% and 1.42\% respectively on these tasks.
\end{abstract}

\section{Introduction}
\label{sec:intro}

The design of early convolutional architectures (CNN) involved choices of hyper-parameters such as: filter size, number of filters at each layer, and padding~\citep{LeCun1998,Krizhevsky2012NIPS}. Since the introduction of  the VGGNet~\citep{Simonyan2014} the design has moved towards following a template: fixed filter size of $3\times3$ and $N$ features maps, down-sample to half the input resolution \textit{only} by the use of either \texttt{maxpool} or strided convolutions~\citep{springenberg2014striving}, and double the number the computed feature maps following each down-sampling operation. This philosophy is used by state of the art models such as ResNet~\citep{He2016ECCV} and DenseNet~\citep{Huang2017densely}. The last two architectures extended the template to include the use of ``skip-connections'' between non-contiguous layers.

Our work extends this template by adding another element, which we refer to as ``coupled ensembling''. In this set-up, the network is decomposed into several branches, each branch being functionally similar to a complete CNN (and able to fully do the classification task alone, generally with a lower performance).
The proposed template achieves performance comparable to state of the art models with a significantly lower parameter count. The proposed modification is simple to implement and we provide a wrapper to compose different standard architectures at: \url{https://github.com/vabh/coupled\_ensembles}.

In this paper, we make the following contributions: (i) we show that given a parameter budjet, splitting a large network into an ensemble of smaller parallel branches of the same type, and jointly training them performs better or at par; (ii) when a final SoftMax (SM) layer is used during the prediction step, we show that ensemble fusion works better when averaging is done before this layer than when it is done after; (iii) when a final Log-Likelihood (LL) layer is used during the training step, we show that ensemble fusion of branches works better when the fusion is done before this layer than when it is done after; (iv) combining all these elements, we significantly improved the performance and/or significantly reduce the parameter count of state-of-the-art neural network architectures on CIFAR and SVHN data sets. (v) We show that such multi-branch networks can be further ensembled at a higher level still producing a significant performance gain.

This paper is organised as follows: in section~\ref{sec:related}, we discuss related work; in section~\ref{sec:ensembles}, we introduce the concept of coupled ensembles and the motivation behind the idea; in section~\ref{sec:expe}, we present the evaluation of the proposed approach and compare it with the state of the art; and we conclude and discuss future work in section~\ref{sec:discussion}.

\section{Related work}
\label{sec:related}

\textbf{Multi-column architectures.} The network architecture that we propose has strong similarities with Cire\c{s}an's Neural Networks Committees~\citep{Ciresan2011ICDAR} and Multi-Column Deep Neural Network (MCDNN)~\citep{Ciresan2012CVPR}, which are a type of ensemble of networks where the ``committee members'' or ``DNN columns'' correspond to our basic block instances (or branches). However, our coupled ensemble networks differ as following: (i) we train a \textit{single} model which is \textit{composed} of sub-networks, while they train each member or column \textit{separately}. 
(ii) we consider a coupled ensemble of smaller networks for improving the performance of a network, given a budget of parameters for the \textit{entire} model. This is contrary to improving it by utilising multiple models of fixed size and therefore multiplying the overall size (though both are not exclusive); 
(iii) we consider placing the averaging (AVG) layer, not only completely at the end of the network but also before the Log-Likelihood (LL) layer during training and before the Soft-Max (SM) layer during training and/or during prediction; (iv) we used the same preprocessing for all branches while they considered different preprocessing (data augmentation) blocks for different members or different subsets of columns; we consider doing this also in the future.

\textbf{Multi-branch architectures.} Multi-branch architectures have been very successful in several vision applications \citep{He2016CVPR,Szegedy2014}. Recently, modifications have been proposed \citep{Xie2016ResNeXt,chollet2016xception} for these architectures using the concept of  ``grouped convolutions'', in order to factorize spatial feature extraction and channel features. These modifications additionally advocate the use of \textit{template} building blocks stacked together to form the complete model. This modification is at the level of the \textit{building blocks} of their corresponding \textit{base} architectures: ResNet and Inception respectively. In contrast we propose a generic modification of the structure of the CNN at the global model level. This includes a template in which the specific architecture of a ``basic block'' is specified, and then this ``basic block'' is replicated as parallel branches to form the final composite model.

To further improve the performance of such architectures, Shake-Shake regularization \citep{Gastaldi17ShakeShake} proposes a stochastic mixture of each of the branches and has achieved good results on the CIFAR datasets. However, the number of epochs required for convergence is much higher compared to the base model. Additionally, the technique seems to depend on the batch size. In contrast, we apply our method using the exact \textit{same} hyper-parameters as used in the underlying CNN.

\cite{Zhao2016connection} investigate the usage of parallel paths in a ResNet, connecting layers to allow information exchange between the paths. However this requires modification at a local level of each of the residual blocks.
In contrast, our method is a generic rearrangement of a given architecture's parameters, which does not introduce additional choices. Additionally, we empirically confirm that our proposed configuration leads to an efficient usage of parameters.

\textbf{Neural network ensembles.} Ensembling is a reliable technique to increase the performance of models for a task. Due to the presence of several local minima, multiple trainings of the exact same neural network architecture can reach a different distribution of errors on a per-class basis. Hence, combining their outputs lead to improved performance on the overall task. This was observed very early \citep{Hansen1990PAMI} and is now commonly used for obtaining top results in classification challenges, despite the increase in training and prediction cost. Our proposed model architecture is not an ensemble of independent networks given that we have a single model made up of parallel branches that is trained. This is similar in spirit to the \texttt{residual block} in ResNet and ResNeXt, and to the \texttt{inception} module in Inception but is done at the full network level. We would like to emphasize here that ``arranging'' a given budget of parameters into parallel branches leads to an increase in performance (as shown in tables~\ref{tab:ensemble_analyze}, \ref{tab:instance_number} and~\ref{tab:leaderboard-single}). Additionally, the classical ensembling approach can still be applied for the fusion of independently trained coupled ensemble networks where it produces a significant performance improvement (as shown in table~\ref{tab:leaderboard-multiple})

\section{Coupled ensembles}
\label{sec:ensembles}
\subsubsection*{Terminology}
For the following discussion, we define some terms:
\begin{itemize}[leftmargin=*]
\item Branch: the proposed model comprises several branches. Each branch takes as input a data point and produces a score vector corresponding to the target classes. Current design of CNNs are referred to as single-branch. The number of branches is denoted by $e$.

\item Basic block: the model architecture used to form a branch. In our experiments, we use DenseNet-BC and ResNet with pre-activation as basic blocks.

\item Fuse Layer: the operation used to combine each of the parallel branches which make up our model. In our experiments, each of the branches are combined by taking the average of each of their predicted log probabilities over the target classes. Section~\ref{sec:expe-combine} explores different choices of operations for the fuse layer.
\end{itemize}

We consider a classification task in which individual samples are always associated to exactly one class, labelled from a finite set. This is the case for CIFAR~\citep{krizhevsky2009CIFAR}, SVHN~\citep{Netzer2011SVHN} and ILSVRC~\citep{Russakovsky2015ILSVRC} tasks. In theory, this should work for other tasks as well (for example, segmentation, object detection, etc.).

We consider neural network models whose last layer outputs a score vector of the same dimension as the number of target classes. This is usually implemented as a linear layer and referred to as a fully connected (FC) layer. This layer can be followed by a  SoftMax (SM) layer to produce a probability distribution over the target classes. During training, this is followed by a loss layer, for example, negative log-likelihood (LL). This is the case for all recent network architectures for image classification\footnote{Though all these networks actually do have a FC layer before the SM one, the last layer need not be a linear layer, as long as it produces one value for each target label. We will refer to the output of each basic block as ``FC''. Similarly, the proposed method may be easily adapted to multi-label classification (i.e. considering non-exclusive classes) by simply replacing the SM (and possibly also the LL) layer(s) by any variant(s) appropriate for multi-label classification. Again, we will refer to their output as  ``SM'' and ``LL''.} \citep{Krizhevsky2012NIPS, Simonyan2014, Szegedy2014, He2016CVPR, Xie2016ResNeXt, Huang2016ECCV, Zagoruyko2016BMVC, Huang2017densely}. The differences among them is related to what is present before the last FC layer. We are agnostic to this internal setup (however complex it may or may not be) because the resulting ``basic block'' always takes an image as input and produces a vector of $N$ values (one for each target class) as output, parametrized by a tensor $W$.

In the case of ensemble predictions, fusion is usually done by computing the individual predictions separately for $e$ model instances and then averaging them. Each of the instances are trained \textit{seperately}. This is functionally equivalent to predicting with a ``super-network'' including the $e$ instances as parallel branches with a final averaging (AVG) layer on top. Such super-networks are generally not  implemented because the branch instances often already correspond to the maximum memory capacity of GPUs. The remaining AVG layer operation can be implemented separately. Alternatively, it is possible to place the averaging layer just after the last FC layer of the basic block instances and before the SM layer, which is then ``factorized''.

In our setup, a model is composed of parallel branches and each branch produces a score vector for the target categories. We explore three options to combine these vectors during training:
\begin{itemize}[leftmargin=*]
\item Activation (FC) average.
\item Probability (SM) average (in practice we do an arithmetic average of log-probabilities).
\item Log Likelihood (LL) average (average of the loss of each branch).
\end{itemize}

We investigate these options in section~\ref{sec:expe-combine}. Note that for inference, averaging the FC layer activations is equivalent to averaging the \texttt{LogSoftMax} values (or to a geometric average of the softmax values, see section~\ref{sec:equiv} of supplementary material).

This transformation of having multiple branches, and combining the branch activations by averaging the log probabilites that they predict for the target categories, leads to a performance improvement, with a \textit{lower} parameter count, in all our experiments (see section~\ref{sec:expe}). The parameter vector $W$ of this composite branched model is the concatenation of the parameter vectors $W_e$ of the $e$ basic blocks with $1 \leq i \leq e$. All parameters are in the ``basic blocks'' as the ``fuse layer'' does not contain any parameters. The $e$ model instances do not really need to share the same architecture.

Three training versions may be considered depending upon whether the ``fuse layer'' is placed after the FC, after the SM or after the LL layer. All the instances are trained simultaneously trough a single loss function and the $W$ parameter vector is obtained in a single training phase. In practice, in the first and the third versions, the SM and LL layers are replaced by a single Cross-Entropy (CE) layer as this is numerically more stable. Similarly, in the second and intermediate case, a \texttt{LogSoftMax} (LSM) layer is used instead of the regular SM layer before the arithmetic averaging (while removing the Log operation in the LL layer). This is equivalent to replacing the arithmetic average by a geometric one with the regular SM and LL layers. The training can be done in four different ways: the three coupled ways correspoding to the ``fuse layer''
plus the the one in which all basic blocks are trained separately in $e$ independent trainings.
Regardless of how the training of the basic blocks has been performed, it is possible to use any of the two coupled ways
for making `coupled prediction' and it is also possible to use the individual basic blocks for making $e$ `individual predictions'.
Not all of these combinations are expected to be consistent or equally efficient but all can be implemented and evaluated as shown in section~\ref{sec:expe-combine}.

\section{Experiments}
\label{sec:expe}

\subsection{Datasets}
\label{sec:datasets}

We evaluate our proposed architecture on the CIFAR \citep{krizhevsky2009CIFAR} and SVHN \citep{Netzer2011SVHN} data sets. CIFAR-10 and CIFAR-100 consist each of 50,000 training images and 10,000 test images, distributed into 10 and 100 categories respectively. SVHN consists of 73,257 training images, 531,131 ``easy'' training images (we use both for training) and 26,032 testing images, distributed into 10 categories. Each image from these datasets is of size 32$\times$32 pixels.

\subsection{Experimental set-up}
\label{sec:exp_setup}

All hyper parameters are set according to the original descriptions of the ``basic block'' that is used. This may not be the optimal setting in our case (especially the learning rate decay schedule) but we do not alter them, so as to not introduce any bias in comparisons. 

For CIFAR-10, CIFAR-100 and SVHN, the input image is normalised by subtracting by the mean image and dividing by the standard deviation. During training on CIFAR datasets, standard data augmentation is used, which comprises random horizontal flips and random crops. For SVHN, no data augmentation is used. However, a dropout ratio of 0.2 is applied in the case of DenseNet when training on SVHN. Testing is done after normalising the input in the same way as during training.

All error rates are given in percentages and correspond to an average computed on the last 10 epochs for CIFAR and on the last 4 epochs for SVHN. This measure is more conservative than the one used by the DenseNet authors (see supplementary material, section~\ref{sec:repro}). For DenseNet-BC, \cite{Amos2017densenet}'s PyTorch implementation has been used. All execution times were measured using a single NVIDIA 1080Ti GPU with the optimal micro-batch\footnote{The micro-bath denotes the number of images samples that were processed in one batch on the GPU so as to have the best throughput. In practice, minibatch (for parameter update) = n$\times$microbatch}. Experiments in section~\ref{sec:branch-indepedent-comparison},~\ref{sec:expe-combine} are done on the CIFAR-100 data set with the ``basic block'' being DenseNet-BC, depth $L=100$, growth rate $k=12$. For experiments in Section~\ref{sec:num_instances}, we consider this same configuration (with a single branch) as our baseline reference point.

\subsection{Comparison with single branch and independent ensembles}
\label{sec:branch-indepedent-comparison}

A natural point of comparison of the proposed branched architecture is with an ensemble of independent models. Rows 2 (coupled training with SM averaging) and 4 (no averaging) in Table~\ref{tab:ensemble_analyze} present the results of these two cases respectively. Row 4 shows the error rate obtained from averaging the predictions of 4 identical models, each of which were trained separately. We see that even though the total number of trainable parameters involved is exactly the same, a \textit{jointly trained branched} configuration gives a much lower test error ($17.61$ vs. $18.42$ with 4 predictions fused at the FC level in both cases).

The next point of comparison is with a \textit{single} branch model comprising a similar number of parameters as the multi branch configuration. The choice of single branch models has been done by: increasing $k$ while keeping $L$ constant, by increasing both $k$ and $L$, or by increasing $L$ while keeping $k$ constant. The last three rows of Table~\ref{tab:ensemble_analyze} show that the error from the multi branch model is considerably lower, as compared to a single branch model ($17.61$ vs. $20.01$, with 4 predictions fused at the FC level).

These observations show that the arranging a given budget of parameters into parallel branches is efficient in terms of parameters, as compared to having a large single branch or multiple independent trainings. In Section~\ref{sec:num_instances} we analyse the relation between the number of branches and the model performance.

\subsection{Choice of Fuse Layer operation}
\label{sec:expe-combine}

In this section, we compare the performance of our proposed branched model for different choices for the positioning of the ``fuse layer'' (see section~\ref{sec:ensembles}). Experiments are carried out to evaluate the best training and prediction fusion combinations. We consider two branched models with $e=2, e=4$, trained in the following conditions: training with fusion after the LL layer, after the SM layer, or after the FC layer.

\begin{table}[htbp]
\begin{center}
\caption{Coupled Ensembles of DenseNet-BCs versus a single model of comparable complexity and study of training / prediction fusion combinations. Top: $e=2$, bottom: $e=4$. Performance is given as the top-1 error rate (mean$\pm$standard deviation for the individual branches) on the CIFAR-100 data set with standard data augmentation. Columns ``$L$'' and ``$k$'' indicate the DenseNet-BC hyper-parameter values of the ``basic block''. Column ``$e$'' indicates the number of branches. Column ``Avg.'' indicates the type of ``fuse layer'' during training: ``none'' for separate trainings (classical ensembling), ``FC'', ``SM'' and ``LL'' for fusion (arithmetic averaging) after the FC, SM and LL layers respectively (not applicable for $e=1$). Column ``Individual'' gives the performance for the individual ``basic blocks'' evaluated separately; Columns ``FC'' and ``SM'' give the performance with fusion (arithmetic averaging) during prediction done after the FC and SM layers respectively. The last three columns give the total number of parameters of the model, the duration of a training iteration (50,000 images) and the prediction time per test image (in batch mode). (*) See supplementary material, section~\ref{sec:repro}; The average and standard deviations are computed here for the independent trainings (comprising 2 and 4 models respectively.).}
\label{tab:ensemble_analyze}
\vspace{2mm}
\begin{tabular}{c c c c c c c c c c}
\hline
$L$ & $k$ & $e$ & Avg. & Individual & FC & SM & Params & Epoch(s) & Test(ms) \\
\hline
100 & 12 & 2 & FC   & {\color{red}{52.68$\pm$22.95}} & 22.25 & {\color{red}{28.78}} & 1.60M & 174 & 0.98 \\
100 & 12 & 2 & SM   & 22.17$\pm$0.32 & {\bf 19.06} & 19.43 & 1.60M & 174 & 0.98 \\
100 & 12 & 2 & LL   & 22.78$\pm$0.08 & 19.33 & 19.91 & 1.60M & 174 & 0.98 \\
\hline
100 & 12 & 2 & none & 23.13$\pm$0.15(*) & 20.44 & 20.44 & 1.60M & 171 & 0.98 \\
\hline
100 & 17 & 1 & n/a  & 21.22$\pm$0.12 & n/a   & n/a   & 1.57M & 121 & 0.67 \\ 
124 & 14 & 1 & n/a  & 21.75$\pm$0.10 & n/a   & n/a   & 1.55M & 135 & 0.77 \\ 
148 & 12 & 1 & n/a  & 20.80$\pm$0.06 & n/a   & n/a   & 1.56M & 159 & 0.90 \\ 
\hline
\vphantom{p} &      &       &       &           &    &    &    &    &    \\ %
\hline
$L$ & $k$ & $e$ & Avg. & Individual & FC & SM & Params & Epoch(s) & Test(ms) \\
\hline
100 & 12 & 4 & FC & {\color{red}{74.36$\pm$26.28}} & 22.55 & {\color{red}{31.92}} & 3.20M & 402 & 2.00 \\
100 & 12 & 4 & SM & 22.29$\pm$0.11 & {\bf 17.61} & 17.68 & 3.20M & 402 & 2.00 \\
100 & 12 & 4 & LL & 22.83$\pm$0.18 & 18.21 & 18.92 & 3.20M & 402 & 2.00 \\
\hline
100 & 12 & 4 & none & 23.13$\pm$0.09(*) & 18.42 & 18.85 & 3.20M & 341 & 2.00 \\
\hline
100 & 25 & 1 & n/a  & 20.61$\pm$0.01 & n/a   & n/a   & 3.34M & 164 & 0.8 \\ 
154 & 17 & 1 & n/a  & 20.02$\pm$0.10 & n/a   & n/a   & 3.29M & 245 & 1.3 \\ 
220 & 12 & 1 & n/a  & 20.01$\pm$0.12 & n/a   & n/a   & 3.15M & 326 & 1.5 \\ 
\hline
\end{tabular}
\end{center}
\end{table}

Table~\ref{tab:ensemble_analyze} shows the performance of the differently trained systems for different prediction configurations: individual average performance of the trained instances (without fusion) and performance of the ensemble system with fusion after the SM layer or after the FC layer. \textit{Note that this table includes models with parameters obtained using different training methods} . We can make the following observations:
\begin{itemize}[leftmargin=*]
\setlength\itemsep{0em}
\item The Avg. FC training with separate predictions (in red) does not work well. This is expected since a similar FC average may be reached with quite unrelated FC instances. The Avg. FC training with Avg. SM prediction (in red) works a bit better but is still not good because the non-linearity of the SM layer distorts the FC average. Indeed, the consistent Avg. FC training with Avg. FC prediction works quite well though it does not yield the best performance.

\item The Avg. FC prediction works at least as well and often significantly better than the Avg. SM prediction whatever the training choice is. This can be explained by the fact that the SM layer compresses values for probabilities close to 0 and 1 while the values remain more spread and transmit more information at the FC layer, even for different training conditions.

\item The average error rate of each of the ``basic blocks'' trained jointly in coupled ensembles with SM fusion is significantly lower than the error rate of the individual instances trained separately. This indicates that the coupling not only forces them to learn complementary features as a group but also to better learn individually. Averaging the log probabilities forces the network to continuously update all branches so as to be consistent with each other. This provides a stronger gradient signal. Additionally, the training loss remains higher compared to single branch models, serving as a regularizer. The error gradient that is back-propagated from the fuse layer is the same for all branches, and this gradient depends on the \textit{combined} predictions. This means that at every step all branches act complementary to the other branches' weight updates. 

\item All ensemble combinations except those based on the Avg. FC training do significantly better than a single network of comparable size and same depth. For a global network size of about 1.6M (resp. 3.2M) parameters, the error rate decreases from 20.80 (resp. 20.01) for the best single instance combination to 19.06 ($-$1.74) with two instances (resp. 17.61 ($-$2.40) with four instances).

\item The best combination seems to be Avg. SM for training with Avg. FC for prediction.

\item The branched model with $e=4$ and Avg. SM for the ``fuse layer'' has the \textit{same} performance as a DenseNet-BC $(L=250, k=24)$ model~\citep{Huang2017densely}, which has about \textit{5 times more} parameters (15.3M versus 3.2M).
\end{itemize}
All the following experiments have Avg. SM for the training ``fuse layer' in the branched models.

\subsection{Choice of the number of branches}
\label{sec:num_instances}

In this section, we investigate the optimal number of branches $e$ for a given model parameter budget. We evaluate on CIFAR-100, with DenseNet-BC as the ``basic block'', and parameter budget equal to $0.8M$ parameters (this is the number of parameters in DenseNet-BC $(L = 100, k = 12)$). Indeed, the optimal number of instances $e$ is likely to depend upon the network architecture, upon the parameter budget and upon the data set but this gives at least one reference. This was investigated again with larger models, and the results are in table~\ref{tab:leaderboard-single} (last four rows).

\begin{table}[htbp]
\begin{center}
\caption{Different number of branches, $e$ for a fixed parameter count. The models are trained  on CIFAR-100 with standard data augmentation. See table~\ref{tab:ensemble_analyze} caption for the meaning of row and column labels. When applicable ($e > 1$), ``fuse layer'' is SM Avg. (*) Average and standard deviation on 10 trials with different seeds; \cite{Huang2017densely} reports 22.27, see supplementary material, section~\ref{sec:repro}.}
\label{tab:instance_number}
\vspace{2mm}
\begin{tabular}{c c c c c c c c c}
\hline
$L$ & $k$ & $e$ & Individual & FC & SM & Params & Epoch(s) & Test(ms) \\
\hline
100 & 12 & 1 & \bf{22.87$\pm$0.17}(*) & n/a & n/a & 800k & 86 & 0.51 \\ %
\hline
 76 & 10 & 2 & 25.58$\pm$0.20 & 21.66 & 22.17 & 720k & 103 & 0.63 \\ 
 88 &  9 & 2 & 25.15$\pm$0.31 & 21.87 & 22.19 & 747k & 119 & 0.71 \\ 
 94 &  8 & 2 & 25.72$\pm$0.20 & 21.95 & 22.22 & 666k & 115 & 0.69 \\ 
100 &  8 & 2 & 25.42$\pm$0.20 & 21.87 & 22.07 & 737k & 126 & 0.75 \\ 
\hline
 70 &  9 & 3 & 26.67$\pm$0.40 & \bf{21.10} & 21.24 & 773k & 129 & 0.77 \\ 
 82 &  8 & 3 & 26.47$\pm$0.17 & \bf{21.25} & 21.46 & 800k & 141 & 0.85 \\ 
 88 &  7 & 3 & 26.92$\pm$0.41 & 22.09 & 22.49 & 698k & 148 & 0.92 \\ 
 94 &  7 & 3 & 26.50$\pm$0.12 & 21.95 & 22.35 & 775k & 160 & 0.98 \\ 
\hline
 64 & 8 & 4  & 28.58$\pm$0.59 & 22.44 & 22.58 & 719k & 142 & 0.88 \\ 
 70 & 8 & 4  & 27.65$\pm$0.48 & 21.50 & 22.12 & 828k & 156 & 0.94 \\ 
\hline
 58 & 7 & 6  & 30.11$\pm$0.53 & 23.87 & 24.22 & 718k & 179 & 1.08 \\ 
 64 & 7 & 6  & 30.65$\pm$0.62 & 23.08 & 23.36 & 840k & 198 & 1.20 \\ 
\hline	
 58 & 6 & 8  & 32.15$\pm$0.00 & 25.95 & 25.70 & 722k & 219 & 1.35 \\ 
 64 & 6 & 8  & 31.52$\pm$0.38 & 24.42 & 24.69 & 843k & 250 & 1.51 \\ 
\hline
\end{tabular}
\end{center}
\end{table}

Table~\ref{tab:instance_number} shows the performance for different configurations of branches $e$, depth $L$, and growth rate $k$. One difficulty is that DenseNet-BC parameter counts are strongly quantified according to the $L$ and $k$ values ($L$ has to be a multiple of 6 modulo 4) and, additionally, to the $e$ value in the coupled ensemble version. This is even more critical in moderate size models like the 800K one targeted here. We selected model configurations with parameters just below the target for making a fair comparison. A few models have slightly more parameters so that some interpolation can be done for possibly more accurate comparisons. We can make the following observations:

\begin{itemize}[leftmargin=*]
\setlength\itemsep{0em}
\item In the considered case (DenseNet-BC, CIFAR-100 and 800K parameters), the optimal number of branches is $e=3,L=70,k=9$. With these parameters, the error rates decreases from 22.87 for the regular $(L = 100, k = 12)$ DenseNet-BC model to 21.10 ($-$1.77).
\item Using 2 to 4 branches yields a significant performance gain over the classical (single branch, $e=1$) case, and even over the original performance of 22.27 reported for the $(L = 100, k = 12)$ DenseNet-BC (see supplementary material, section~\ref{sec:repro}).
\item Using 6 or 8 branches performs significantly less well.
\item Slightly varying the $L$, $k$ and $e$ hyper-parameters around their optimal value does not lead to a significant performance drop, showing that the coupled ensemble approach and the DenseNet-BC architecture are quite robust relatively to these choices.
\item The gain in performance comes at the expense of an increased training and prediction times even though the model size does not change. This is due to the use of smaller values of $k$ that prevents good parallelism efficiency. This increase is relatively smaller with bigger networks.
\end{itemize}
The same experiment was done on a validation set with a 40k/10k random split of the CIFAR-100 training set and we could draw the same conclusions from there; they led to predict that the $(L = 82, k = 8, e = 3)$ combination should be the best one on the test set. The $(L = 70, k = 9, e = 3)$ combination appears to be slightly better here but the difference is probably not statistically significant.

\subsection{Comparison with the state of the art}
\label{sec:leaderboard-single}

We have evaluated the coupled ensemble network approach with networks of various sizes. We used again \cite{Huang2017densely}'s DenseNet-BC architecture as the ``basic block'' since this was the current state of the art or very close to it at the time we started these experiments. We used \cite{Amos2017densenet}'s PyTorch DenseNet-BC implementation both for the multi-branch (coupled ensembles) and single-branch (classical) experiments. We also evaluated the approach using \cite{He2016ECCV}'s ResNet with pre-activation as the basic block to check if the coupled ensemble approach works well with other architectures.

Table~\ref{tab:leaderboard-single} reports in the upper part the results obtained by the current best systems (see section~\ref{sec:related} for references) and the results obtained with our coupled ensembles approach in the lower part. All results presented in this table correspond to the training of a single, possibly big, network. Even if ensembles are considered, they are always coupled as described in section~\ref{sec:ensembles} and trained as a single global network. A further level of ensembling involving multiple trainings is considered in section~\ref{sec:leaderboard-multiple}. Results are presented on the CIFAR 10 and 100 data set with standard data augmentation and on SVHN using the extra training data.

\begin{table}[htbp]
\begin{center}
\caption{Classification error comparison with the state of the art, for a single model.} 
\label{tab:leaderboard-single}
\begin{tabular}{l c c c c}
\hline
System & C10+ & C100+ & SVHN  & \#Params \\
\hline
ResNet $L=110$ $k=64$                           & 6.61 &   -   &  -   &  1.7M \\
\hline
ResNet stochastic depth $L=110$ $k=64$          & 5.25 & 24.98 &  -   &  1.7M \\
ResNet stochastic depth $L=1202$ $k=64$         & 4.91 &   -   &  -   & 10.2M \\
\hline
ResNet pre-activation $L=164$ $k=64$            & 5.46 & 24.33 &  -   &  1.7M \\
ResNet pre-activation $L=1001$ $k=64$           & 4.92 & 22.71 &  -   & 10.2M \\
\hline
DenseNet    $L=100$ $k=24$                      & 3.74 & 19.25 & 1.59 & 27.2M \\
\hline
DenseNet-BC $L=100$ $k=12$ (Torch)              & 4.51 & 22.27 & 1.76 & 0.80M \\
DenseNet-BC $L=250$ $k=24$ (Torch)              & 3.62 & 17.60 &  -   & 15.3M \\
DenseNet-BC $L=190$ $k=40$ (Torch)              & 3.46 & 17.18 &  -   & 25.6M \\
\hline
Shake-Shake C10 Model S-S-I                     & 2.86 &   -   &  -   & 26.2M \\
Shake-Shake C100 Model S-E-I                    &  -   & 15.85 &  -   & 34.4M \\
\hline
Snapshot Ensemble DenseNet-40 ($\alpha_0=0.1$)  & 4.99 & 23.34 & 1.64 &  6.0M \\ 
Snapshot Ensemble DenseNet-40 ($\alpha_0=0.2$)  & 4.84 & 21.93 & 1.73 &  6.0M \\ 
Snapshot Ensemble DenseNet-100 ($\alpha_0=0.2$) & 3.44 & 17.41 &  -   &  163M \\ 
\hline
SGDR WRN-28-10 \cite{Loshchilov2017ICLR}        & 4.03 & 19.57 &  -   & 36.5M \\
SGDR WRN-28-10 3 snapshots                      & 3.51 & 17.75 &  -   &  110M \\ 
\hline
ResNeXt-29, 8$\times$64d \cite{Xie2016ResNeXt}  & 3.65 & 17.77 &  -   & 34.4M \\
ResNeXt-29, 16$\times$64d \cite{Xie2016ResNeXt} & 3.58 & 17.31 &  -   & 68.1M \\
\hline
DFN-MR2 \cite{Zhao2016connection}               & 3.94 & 19.25 & 1.51 & 14.9M \\
DFN-MR3 \cite{Zhao2016connection}               & 3.57 & 19.00 & 1.55 & 24.8M \\
\hline
IGC-L450M2 \cite{ZhangInterleaved}              & 3.25 & 19.25 &  -   & 19.3M \\
IGC-L32M26 \cite{ZhangInterleaved}              & 3.31 & 18.75 & 1.56 & 24.1M \\
\hline
\hline
ResNet pre-activation $L=65$  $k=64$ $e=2$      & 5.26 & 23.24 &  -   &  1.4M \\
ResNet pre-activation $L=164$ $k=64$ $e=2$      & 4.24 & 19.92 &  -   &  3.4M \\
ResNet pre-activation $L=164$ $k=64$ $e=4$      & 3.96 & 18.84 &  -   &  6.8M \\
\hline 
DenseNet-BC $L=100$ $k=12$ $e=1$                & 4.77 & 22.87 & 1.79 &  0.8M \\
DenseNet-BC $L=112$ $k=16$ $e=1$                & 4.47 & 20.73 & 1.83 &  1.7M \\
DenseNet-BC $L=130$ $k=20$ $e=1$                & 4.06 & 19.03 & 1.84 &  3.4M \\
DenseNet-BC $L=160$ $k=24$ $e=1$                & 3.98 & 18.92 & 1.88 &  6.9M \\
DenseNet-BC $L=166$ $k=32$ $e=1$                & 4.03 & 20.03 & 1.88 & 13.0M \\
DenseNet-BC $L=190$ $k=40$ $e=1$                & 4.04 & 18.19 & 1.79 & 25.8M \\
\hline
DenseNet-BC $L=82$  $k=8$  $e=3$                & 4.30 & 21.25 & 1.66&  0.8M \\
DenseNet-BC $L=82$ $k=10$ $e=4$                 & 3.78 & 19.92 & 1.62 & 1.6M \\
DenseNet-BC $L=88$ $k=14$ $e=4$                 & 3.57 & 17.68 & 1.55 & 3.5M \\
DenseNet-BC $L=88$ $k=20$ $e=4$                 & 3.18 & 16.79 & 1.57 & 7.0M \\
DenseNet-BC $L=94$ $k=26$ $e=4$                 & 3.01 & 16.24 & 1.50 & 13.0M \\
\hline
DenseNet-BC $L=118$ $k=35$ $e=3$                & 2.99 & 16.18 & 1.50 & 25.7M \\
DenseNet-BC $L=106$ $k=33$ $e=4$                & 2.99 & 15.68 & 1.53 & 25.1M \\
DenseNet-BC $L=76$  $k=33$ $e=6$                & 2.92 & 15.76 & 1.50 & 24.6M \\
DenseNet-BC $L=64$  $k=35$ $e=8$                & 3.13 & 15.95 & 1.50 & 24.9M \\
\hline
\end{tabular}
\end{center}
\end{table}

For the ResNet \texttt{pre-act} architecture, the ensemble versions with 2 or 4 branches leads to a significantly better performance than single branch versions with comparable or higher number of parameters.

Regarding the DenseNet-BC architecture, we considered 6 different network sizes, roughly following multiples of powers of 2 and ranging from 0.8M up to 25.6M parameters, with two extremes corresponding to those for which the error rates were available~\cite{Huang2017densely}. We chose these values for the depth $L$ and growth rate $k$ for these points and interpolated between them according to a log scale as much as possible. Our experiments showed that the trade-off between $L$ and $k$ is not critical for a given overall parameter count. This was also the case for choosing between the number of branches $e$, depth $L$ and growth rate, $k$ for a given overall parameter count budget as long as $e \geq 3$ (or even $e \geq 2$ for small networks). For the 6 configurations, we experimented with both the single-branch (classical) and multi-branch versions of the model, with $e=4$. Additionally, for the largest model, we tried $e=3,6,8$ branches.

For the single branch version with extreme network sizes, we obtained error rates significantly lower than those reported by \cite{Huang2017densely}. From what we have checked, the Lua Torch implementation they used and the PyTorch one we used are equivalent. The difference may be due to the fact that we used a more conservative measure of the error rate (on the last iterations) and from statistical differences due to different initializations and/or to non-deterministic computations (see section~\ref{sec:repro} in supplementary material). Still, the coupled ensemble version leads to a significantly better performance for all network sizes, even when compared to DenseNet-BC's reported performance.

Our larger models of coupled DenseNet-BCs (error rates of 2.92\% on CIFAR 10, 15.68\% on CIFAR 100 and 1.50\% on SVHN) perform better than or are on par with all current state of the art implementations that we are aware of at the time of publication of this work. Only the Shake-Shake S-S-I model~\citep{Gastaldi17ShakeShake} performs slightly better on CIFAR 10.

We also compare the performance of coupled ensembles with model architectures that were `learnt' in a meta learning scenario. The results are presented in the supplementary material, section~\ref{sec:sup-learnt}.

\subsection{Ensembles of coupled ensembles}
\label{sec:leaderboard-multiple}

The coupled ensemble approach is limited by the size of the network that can fit into GPU memory during the training and the time that such training takes. With the hardware we have access to, it was not possible to go much beyond the 25M-parameter scale. For going further, we resorted to the classical ensembling approach based on independent trainings. An interesting question was whether we could still significantly improve the performance since the classical approach generally plateaus after quite a small number of models and the coupled ensemble approach already include several. For instance the SGDR with snapshots approach \citep{Loshchilov2017ICLR} has a significant improvement from 1 to 3 models but not much improvement from 3 to 16 models (see tables~\ref{tab:leaderboard-single} and~\ref{tab:leaderboard-multiple}). As doing multiple times the same training is quite costly when models are large, we instead ensembled the four large coupled ensemble models that we trained for the four values of $e=3, 4, 6, 8$. Results are shown in table~\ref{tab:leaderboard-multiple}. We obtained a significant gain by fusing two models and a quite small one from any further fusion of three or four of them. To the best of our knowledge, these ensembles of coupled ensemble networks outperform all state of the art implementations including other ensemble-based ones at the time of publication of this work.

\begin{table}[htbp]
\begin{center}
\caption{Classification error comparison with the state of the art, multiple model trainings.} 
\label{tab:leaderboard-multiple}
\begin{tabular}{l c c c c}
\hline
System & C10+ & C100+ & SVHN  & \#Params \\
\hline
SGDR WRN-28-10  3 runs $\times$ 3 snapshots        & 3.25 & 16.64 &  -   &  329M \\ 
SGDR WRN-28-10 16 runs $\times$ 3 snapshots        & 3.14 & 16.21 &  -   & 1752M \\ 
\hline
DenseNet-BC ensemble of ensembles $e=6,4$          & 2.72 & 15.13 & 1.42 & 50M \\
DenseNet-BC ensemble of ensembles $e=6,4,3$        & 2.68 & 15.04 & 1.42 & 75M \\
DenseNet-BC ensemble of ensembles $e=8,6,4,3$      & 2.73 & 15.05 & 1.41 & 100M \\
\hline
\end{tabular}
\end{center}
\end{table}

\subsection{Parameter Usage}

\begin{figure}[htbp]
\begin{center}
\includegraphics[width=0.92\linewidth]{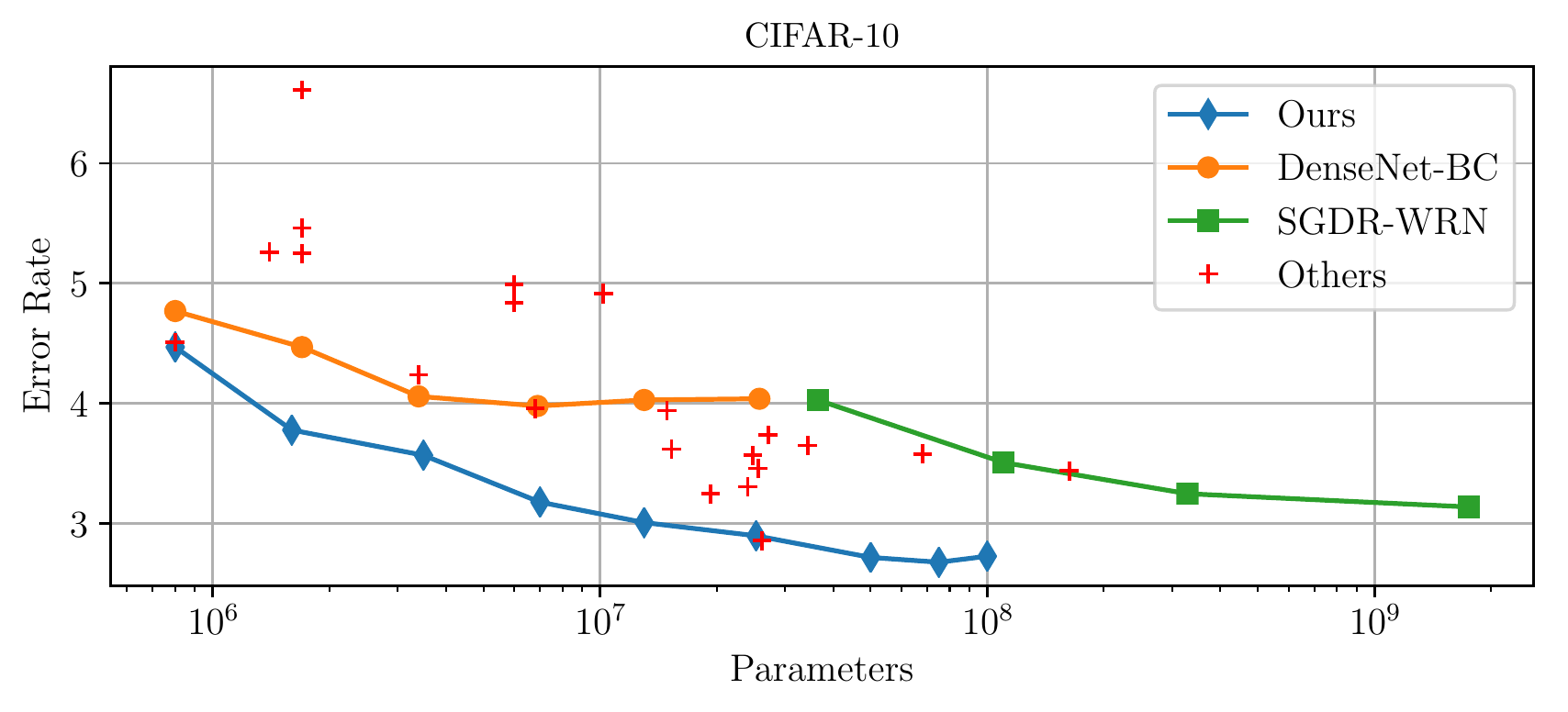}
\includegraphics[width=0.92\linewidth]{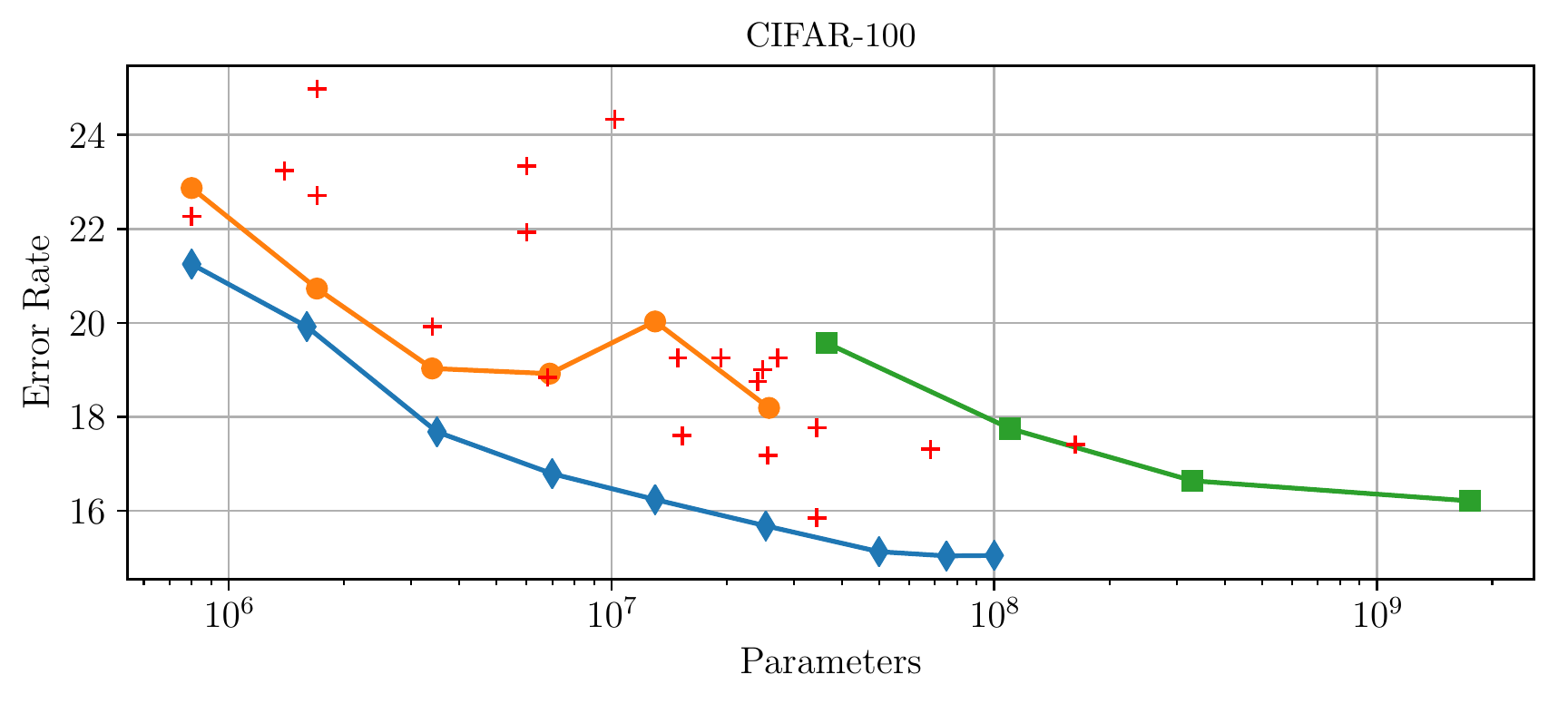}
\includegraphics[width=0.92\linewidth]{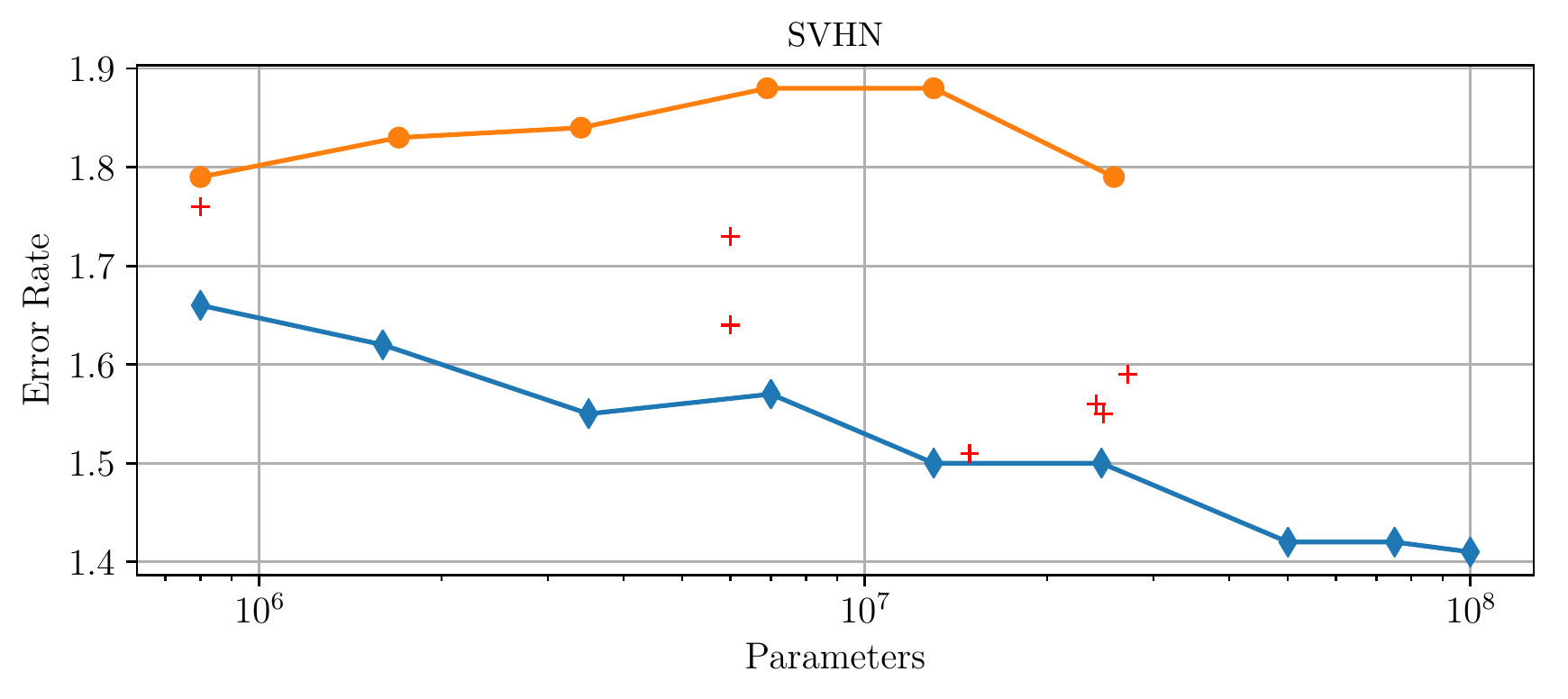}
\caption{Comparison of parameter usage between different methods. Results on  CIFAR-10 (top), CIFAR-100 (middle) and SVHN (bottom) tasks. ``DenseNet-BC'': single-branch (PyTorch) configuration; ``Ours'': proposed multiple branch configuration using DenseNet-BC as the basic block, single training up to 25M parameters and ensembles of coupled ensembles beyond; SGDR-WRN: snapshot ensembles with SGDR on Wide ResNets up to 110M parameters and ensembles of snapshot ensembles beyond; "Other": all other architectures mentioned in tables~\ref{tab:leaderboard-single} and~\ref{tab:leaderboard-multiple}.}
\label{fig:param_usage}
\end{center}
\end{figure}

Figure~\ref{fig:param_usage} shows that for a given parameter count, coupled ensemble networks or ensembles of coupled ensemble networks perform significantly better for all parameter budgets. Our approach is shown by the blue diamonds.

\section{Discussion}
\label{sec:discussion}

The proposed approach consists in replacing a single deep convolutional network by a number of ``basic blocks'' which resemble standalone CNN models. The intermediate score vectors produced by each of the basics blocks are coupled via a ``fuse layer''. At training time, this is done by taking an arithmetic average of their log-probabilites for the targets. At test time the score vectors are averaged following the output from each score vector. Both of these aspects leads to a significant performance improvement over a single branch configuration. This improvement comes at the cost of a small increase in the training and prediction times. The proposed approach leads to the best performance for a given parameter budget as can be seen in tables~\ref{tab:leaderboard-single} and~\ref{tab:leaderboard-multiple}, and in figure~\ref{fig:param_usage}.

The increase in training and prediction times is mostly due to the sequential processing of branches during the forward and backward passes. The smaller size of the branches makes the data parallelism on GPUs less efficient. This effect is not as pronounced for larger models. This could be solved in two ways. First, as there is no data dependency between the branches (before the averaging layer) it is possible to extend the data parallelism to the branches, restoring the initial level of parallelism. This can be done by implementing a parallel implementation of multiples 2D convolutions at the same time. Second or alternatively, when multiple GPUs are used, it is possible to spread the branches over the GPUs.

We have currently evaluated the coupled ensemble approach only on relatively small data sets. We therefore plan to conduct experiments on ImageNet~\citep{Russakovsky2015ILSVRC} to check whether it will work equally well on large collections.

\subsubsection*{Acknowledgements}

This work has been partially supported by the LabEx PERSYVAL-Lab (ANR-11-LABX-0025-01).
Experiments presented in this paper were partly carried out using the Grid'5000 test-bed, supported by a scientific interest group hosted by Inria and including CNRS, RENATER and several Universities as well as other organizations (see https://www.grid5000.fr).

\bibliography{iclr18}
\bibliographystyle{apalike}

\newpage
\appendix
{\Large Supplementary Material}

\section{Implementation}
\label{sec:implem}

\begin{figure}[htbp]
\begin{center}
\includegraphics[width=0.8\linewidth]{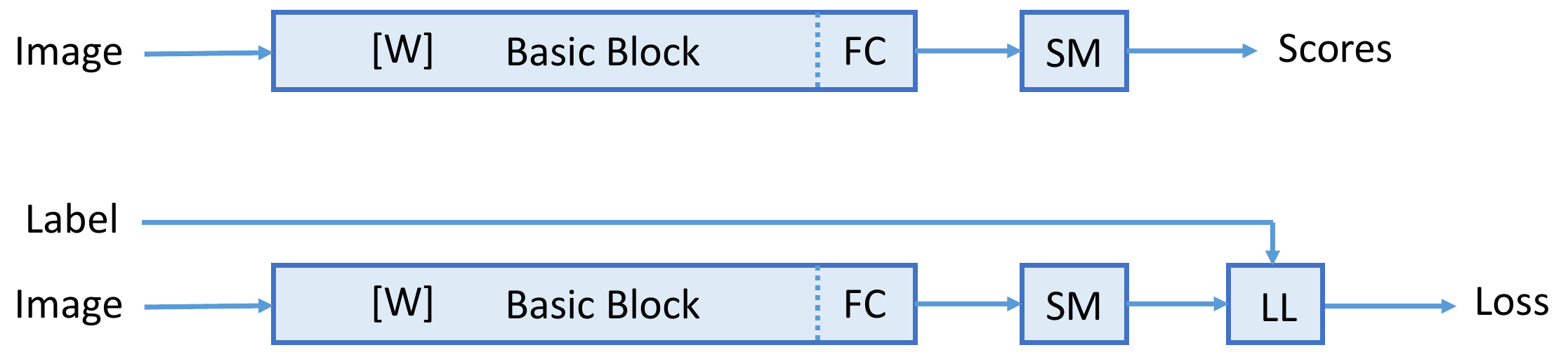}
\caption{Versions of the base network. Top: test, bottom: train.}
\label{fig:base_network}
\end{center}
\end{figure}

\begin{figure}[htbp]
\begin{center}
\includegraphics[width=0.85\linewidth]{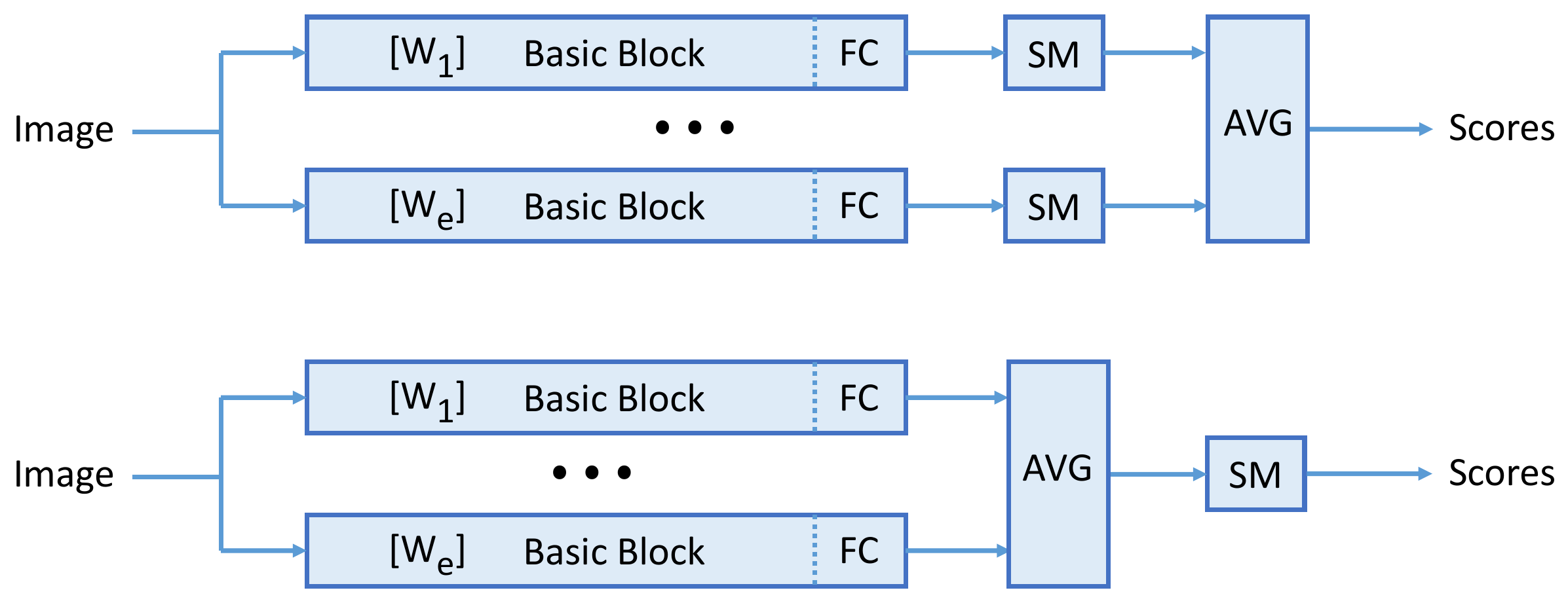}
\caption{Test versions of coupled ensemble networks. Top: SM (classical) fusion, bottom: FC fusion. AVG: averaging layer.}
\label{fig:ensemble_test}
\end{center}
\end{figure}

Figure~\ref{fig:base_network} shows the common structure of the test (top) and train (bottom) versions of networks used as basic blocks. Figure~\ref{fig:ensemble_test} shows how it is possible to place the averaging layer just after the last FC layer of the basic block instances and before the SM layer which is then ``factorized''. The $e$ model instances do not need to share the same architecture. Figure~\ref{fig:ensemble_test} shows how it is possible to place the averaging layer just after the last FC layer, just after the SM (actually LSM (LogSoftMax), which is equivalent to do a geometric mean of the SM values) layer, or just after the LL layer.

\begin{figure}[htbp]
\begin{center}
\includegraphics[width=0.92\linewidth]{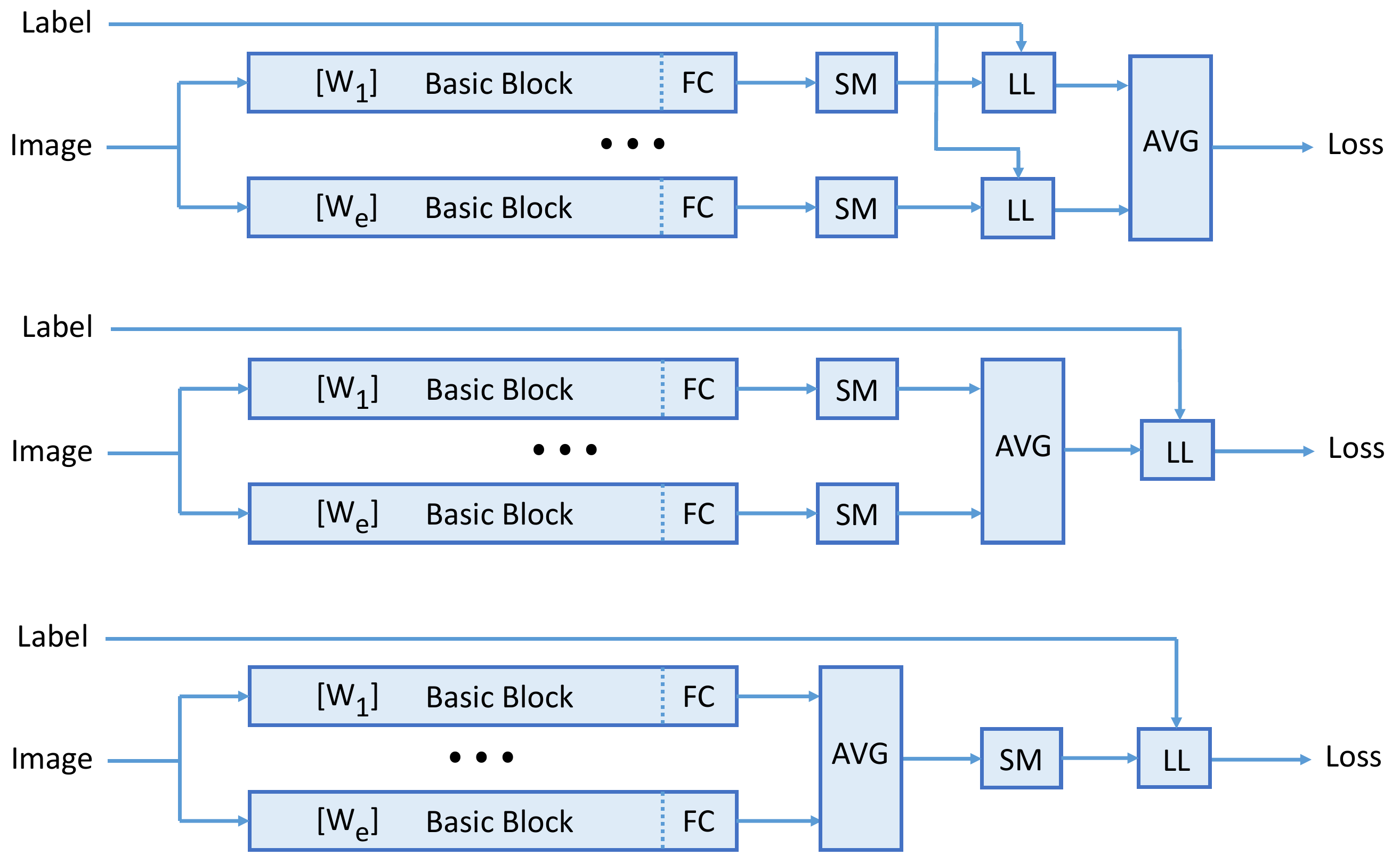}
\caption{Train versions of coupled ensemble networks. Top: LL fusion, middle: SM fusion, bottom: FC fusion.}
\label{fig:ensemble_train}
\end{center}
\end{figure}

We reuse ``basic blocks'' from other groups (with appropriate credits, please let us know if any is missing or requires updating) in their original form as much as possible both for efficiency and for ensuring more meaningful comparisons. When available, we use the PyTorch implementations.

Each of the $e$ branches is defined by a parameter vector $W_e$ containing the same parameters as the original implementation. The global network is defined by a parameter vector $W$ which is a concatenation of all the $W_e$ parameter vectors. When training is done in the coupled mode and the prediction is done in a separate mode or vice-versa, a dedicated script is used for splitting the $W$ vector into the $W_e$ ones or vice-versa. In all coupled networks, for all train versions and for all test version, the same global parameter vector $W$ is used with the same split and defining the same basic block functions. This is how we can combine in any way all of the four possible training conditions with all the three possible prediction conditions, even though not all of them are consistent or equally efficient.

The overall network architecture is determined by:
\begin{itemize}[leftmargin=*]
\item the global hyper-parameter specifying the train versus test mode;
\item the global hyper-parameter $e$ specifying the number of branches;
\item the global hyper-parameter specifying after which layer the AVG layer should be placed (FC, SM or LL);
\item either one basic block to be replicated $e$ times with its own hyper-parameters or a list of $e$ basic blocks, each with its own hyper-parameters.
\end{itemize}

\section{Test time equivalence between FC average and LogSoftMax average}
\label{sec:equiv}

Given branches $E={E_1, E_2, .. E_e}$, each $E_i$ produces a score vector of dimension $C$, where $C$ is the number of categories. An element of $E_i$ is referenced as $E_i^c$, were $c \in [1, C]$. FC\_Average denotes averaging the raw activations from each branch. LSM\_Average denotes averaging across branches, after a LogSoftMax operation in applied on each branch activation vector, separately.

Case 1: FC\_average: $Scores_{FC}^c = \sum_{i=1}^e E_i^c$

Case 2: 
\begin{align*}
LogSoftMax(E_n^c) &= \log \frac{\exp(E_e^c)}{\sum_c \exp(E_e^c)} \\
					&= \log \exp(E_e^c) - \log \sum_c \exp(E_e^c) \\
                    & = E_e^c - Z_e
\end{align*}

LSM\_average: $Scores_{LSM}^c = \sum_{i=1}^e E_i^c - \sum_{i=1}^e Z_i$, where $Z_e = \log \sum_C \exp(E_e^c)$.
Hence we see that the LSM\_average score vector is a translated version of the FC\_average score vector. Also, doing an arithmetic average of LogSoftMax values is equivalent to doing a geometric average of SoftMax values. This holds during inference where we are interested only in the maximum value.

\section{Comparison with \textit{Learnt} architectures}\label{sec:sup-learnt}

In table~\ref{tab:leaderboard-learnt}, we compare the parameter usage and performance of the branched coupled ensembles with model architectures that were recovered using meta learning techniques.

\begin{table}[htbp]
\begin{center}
\caption{Classification error comparison with \textit{learnt} architectures.} 
\label{tab:leaderboard-learnt}
\begin{tabular}{l c c c c}
\hline
System & C10+ & C100+ & SVHN  & \#Params \\
\hline
Neural Architecture Search v3~\citep{zoph2016neural} & 3.65 & - &  -   & 37.4M \\
NASNet-A~\citep{zoph2017learning}                    & 3.41 & - &  -   & 3.3M \\ 
\hline
DenseNet-BC $L=82$ $k=10$ $e=4$                 & 3.78 & 19.92 & 1.62 & 1.6M \\
DenseNet-BC $L=88$ $k=14$ $e=4$                 & 3.57 & 17.68 & 1.55 & 3.5M \\
DenseNet-BC $L=88$ $k=20$ $e=4$                 & 3.18 & 16.79 & 1.57 & 7.0M \\
\hline
\end{tabular}
\end{center}
\end{table}

\section{Performance measurement and reproducibility issues\label{sec:repro}}

When attempting to compare the relative performance of different methods, we face the issue of the reproducibility of the experiments and of the statistical significance of the observed difference between performance measures. Even for a same experiment, we identified the five following sources of variation in the performance measure:
\begin{itemize}[leftmargin=*]
\item Underlying framework for the implementation: we made experiments with Torch7 (lua) and with PyTorch.
\item Random seed for the network initialization.
\item CuDNN non-determinism during training: GPU associative operations are by default fast but non-deterministic. We observed that the results varies even for a same tool and the same seed. In practice, the observed variation is as important as when changing the seed.
\item Fluctuations associated to the computed moving average and standard deviation in batch normalization: these fluctuations can be observed even when training with the learning rate, the SGD momentum and the weight decay all set to $0$. During the last few epochs of training, their level of influence is the same as with the default value of these hyper-parameters.
\item Choice of the model instance chosen from training epochs: the model obtained after the last epoch, or the best performing model. Note that choosing the best performing model involves looking at test data.
\end{itemize}

Regardless of the implementation, the numerical determinism, the Batch Norm moving average, and the epoch sampling questions, we should still expect a dispersion of the evaluation measure according to the choice of the random initialization since different random seeds will likely lead to different local minima. It is generally considered that the local minima obtained with ``properly designed and trained'' neural networks should all have similar performance \citep{kawaguchi2016deep}. We do observe a relatively small dispersion (quantified by the standard deviation below) confirming this hypothesis. This dispersion may be small but it is not negligible and it complicates the comparisons between methods since differences in measures lower than their dispersions is likely to be non-significant. Classical statistical significance tests do not help much here since differences that are statistically significant in this sense can be observed between models obtained just with different seeds (and even with the same seed), everything else being kept equal.

Experiments reported in this section gives an estimation of the dispersion in the particular case of a moderate scale model. We generally cannot afford doing a large number of trials for larger models.

We tried to quantify the relative importance of the different effects in the particular case of DenseNet-BC with $L=100, k=12$ on CIFAR 100. Table~\ref{tab:perf-repro} shows the results obtained for the same experiment in the four groups of three rows. We tried four combinations corresponding to the use of Torch7 versus PyTorch and to the use of the same seed versus the use of different seeds. For each of these configuration, we used as the performance measure: (i) the error rate of the model computed at the last epoch or (ii) the average of the error rate of the models computed at the last 10 epochs, (iii) the error rate of the model having the lowest error rate over all epochs. For these $2\times2\times3$ cases, we present the minimum, the median, the maximum and the mean$\pm$standard deviation over 10 measures corresponding to 10 identical runs (except for the seed when indicated). Additionally, in the case of the average of the error rate of the models computed at the 10 last epochs, we present the root mean square of the standard deviation of the fluctuations on the last 10 epochs (which is the same as the square root of the mean of their variance). We make the following observations:
\begin{itemize}[leftmargin=*]
\item There does not seem to be a significant difference between Torch7 and PyTorch implementations;
\item There does not seem to be a significant difference between using a same seed and using different seeds; the dispersion observed using the same seed (with everything else being equal) implies that there is no way to exactly reproduce results;
\item There does not seem to be a significant difference between the means over the 10 measures computed on the single last epoch and the means over the 10 measures computed on the last 10 epochs;
\item The standard deviation of the measures computed on the 10 runs is slightly but consistently smaller when the measures are computed on the last 10 epochs than when they are computed on the single last epoch; this is the same for the difference between the best and the worst measures; this was expected since averaging the measure on the last 10 epochs reduces the fluctuations due to the moving average and standard deviation computed in batch normalization and possibly too the the random fluctuations due to the final learning steps;
\item The mean of the measures computed on the 10 runs is significantly lower when the measure is taken at the best epoch than when they are computed either on the single last epoch or on the last 10 epochs. This is expected since the minimum is always below the average. However, presenting this measure involves using the test data for selecting the best model. 
\end{itemize}

\begin{table}[htbp]
\begin{center}
\caption{Performance measurement and reproducibility issues. Statistics on 10 runs.}
\label{tab:perf-repro}
\begin{tabular}{ccccccccccc}
\hline
Seeds & Impl. & Last & $L$ & $k$ & $e$ & Min. & Med. & Max. & Mean$\pm$SD & RMS(SD) \\
\hline
diff. &  PyT. &   1  & 100 & 12 & 1 & 22.64 & 22.80 & 23.22 & 22.89$\pm$0.21 &  n/a   \\
diff. &  PyT. &  10  & 100 & 12 & 1 & 22.67 & 22.83 & 23.14 & 22.87$\pm$0.17 &  0.13  \\
diff. &  PyT. & best & 100 & 12 & 1 & 22.13 & 22.56 & 22.91 & 22.54$\pm$0.24 &  n/a   \\
\hline
same  &  PyT. &   1  & 100 & 12 & 1 & 22.77 & 23.05 & 23.55 & 23.06$\pm$0.23 &  n/a   \\
same  &  PyT. &  10  & 100 & 12 & 1 & 22.81 & 22.98 & 23.49 & 23.04$\pm$0.22 &  0.11  \\
same  &  PyT. & best & 100 & 12 & 1 & 22.44 & 22.67 & 23.02 & 22.71$\pm$0.18 &  n/a   \\
\hline
diff. & LuaT. &   1  & 100 & 12 & 1 & 22.55 & 22.94 & 23.11 & 22.90$\pm$0.20 &  n/a   \\
diff. & LuaT. &  10  & 100 & 12 & 1 & 22.55 & 22.89 & 23.08 & 22.86$\pm$0.20 &  0.12  \\
diff. & LuaT. & best & 100 & 12 & 1 & 22.17 & 22.52 & 22.75 & 22.49$\pm$0.18 &  n/a   \\
\hline
same  & LuaT. &   1  & 100 & 12 & 1 & 22.33 & 22.82 & 23.58 & 22.82$\pm$0.34 &  n/a   \\
same  & LuaT. &  10  & 100 & 12 & 1 & 22.47 & 22.92 & 23.51 & 22.87$\pm$0.30 &  0.12  \\
same  & LuaT. & best & 100 & 12 & 1 & 22.24 & 22.51 & 23.24 & 22.54$\pm$0.29 &  n/a   \\
\hline
\hline
diff. &  PyT. &   1  &  82 &  8 & 3 & 21.27 & 21.44 & 21.70 & 21.49$\pm$0.15 &  n/a   \\
diff. &  PyT. &  10  &  82 &  8 & 3 & 21.24 & 21.46 & 21.63 & 21.45$\pm$0.11 &  0.12  \\
diff. &  PyT. & best &  82 &  8 & 3 & 20.84 & 21.18 & 21.30 & 21.14$\pm$0.14 &  n/a   \\
\hline
\hline
diff. &  PyT. &   1  & 100 & 12 & 4 & 17.24 & 17.71 & 17.86 & 17.65$\pm$0.18 &  n/a   \\
diff. &  PyT. &  10  & 100 & 12 & 4 & 17.37 & 17.67 & 17.81 & 17.66$\pm$0.14 &  0.11  \\
diff. &  PyT. & best & 100 & 12 & 4 & 17.11 & 17.46 & 17.66 & 17.45$\pm$0.16 &  n/a   \\
\hline
\end{tabular}
\end{center}
\end{table}

Following these observations, we propose a method for ensuring the best reproducibility and the fairest comparisons. Choosing the measure as the minimum of the error rate for all models computed during the training seems neither realistic nor a good practice since we have no way to know which model will be the best one without looking at the results (cross-validation cannot be used for that) and this is like tuning on the test set. Even though this is not necessarily unfair for system comparison if the measures are done in this condition for all systems, this does introduce a bias for the absolute performance estimation. Using the error rate at the last iteration or at the 10 last iteration does not seem to make a difference in the mean but the standard deviation is smaller for the latter, therefore this one should be preferred when a single experiment is conducted. We also checked that using the 10 or the 25 last epochs does not make much difference (learning at this point does not seem to lead to further improvement). A value different from 10 can be used and this is not critical. In all the CIFAR experiments reported in this paper, we used the average of the error rate for the models obtained at the last 10 epochs as this should be (slightly) more robust and more conservative. The case for SVHN experiments is slightly different since there is a much smaller number of much bigger epochs; we used the last 4 iterations in this case.

These observations have been made in a quite specific case but the principle and the conclusions (use of the average of the error rate from the last epochs should lead to more robust and conservative results) are likely to be general. Table~\ref{tab:perf-repro} also shows the results for a coupled ensemble network of comparable size and for a coupled ensemble network four times bigger. Similar observations can be made and, additionally, we can observe that both the range and the standard deviations are smaller. This might be because an averaging is already made between the branches leading to a reduction of the variance. Though this requires confirmation at larger scales, coupled ensemble networks might lead to both a better and a more stable performance.



\end{document}